\documentclass[11pt]{article}
\usepackage{acl2015}
\usepackage{times}
\usepackage{url}
\usepackage{latexsym}
\usepackage{multirow}
\usepackage{subcaption}
\usepackage{pbox}
\usepackage{array}
\usepackage{graphicx}
\usepackage{algorithm}
\usepackage{xcolor,colortbl}
\usepackage[noend]{algpseudocode}

\newcommand{\swallow}[1]{}

\newcommand{\fullversion}[1]{}

\makeatletter
\def\BState{\State\hskip-\ALG@thistlm}
\makeatother

\title{False-Friend Detection and Entity Matching via Unsupervised Transliteration}

\author{Yanqing Chen\\
             Stony Brook University\\
            100 Nicolls Road\\
             Stony Brook, NY 11790, USA\\
             {\tt cyanqing@cs.stonybrook.edu}
  \And
	Steven Skiena\\
          Stony Brook University\\
          100 Nicolls Road\\
          Stony Brook, NY 11790, USA\\
          {\tt skiena@cs.stonybrook.edu}}

\date{}

\begin{document}
\maketitle

\begin{abstract}

Transliterations play an important role in multilingual entity reference resolution, because proper names increasingly
travel between languages in news and social media.
Previous work associated with machine translation targets transliteration only single between language pairs,
focuses on specific classes of entities (such as cities and celebrities) and relies on manual curation,
which limits the expression power of transliteration in multilingual environment.

\swallow{
As a result of globalization, fresh words will sometimes appear as Out Of Vocabulary words (OOV) with multiple possible transliterations in foreign languages. Constructing a multilingual transliteration model is both interesting and challenging. It provides opportunities to learn connections between character and pronunciations across different languages.
}

By contrast, we present an unsupervised transliteration model covering 69 major languages that can generate good transliterations for arbitrary strings between any language pair.
Our model yields top-(1, 20, 100) averages of (32.85\%, 60.44\%, 83.20\%) in matching gold standard transliteration compared to results from a recently-published system of (26.71\%, 50.27\%, 72.79\%).
We also show the quality of our model in detecting true and false friends from Wikipedia high frequency lexicons. Our method indicates a strong signal of pronunciation similarity and boosts the probability of finding true friends in 68 out of 69 languages.

\end{abstract}

\section{Introduction}

Transliterations play an important role in multilingual entity reference resolution, because proper names increasingly travel between languages. This process tends to create a substantial number of out of vocabulary (OOV) words in the multilingual analysis of news and social media.
When ``Gangnam style" topped the music charts of more than 30 countries, a word imported from Korean suddenly became part of the language spoken by millions of people around the world. News events like the catastrophic failure at nuclear power plant bring words associated with new people and places (`Fukushima") across languages into common use.
These words do not reside in standard vocabulary lexicons, but are generated as needed via a process of transliteration.
Detecting transliterated word pairs contributes to many language processing tasks, including entity resolution, translation, topic classification and sentiment analysis, as well as facilitates studying linguistic phenomenon like cross-language morphologic evolution. 
However, previous transliteration systems generally focus on a small number of language pairs.
Further, they only consider morphological similarity even in translation systems, creating a problem of "false friends" of word pairs which look/sound alike but mean different things.

\begin{figure}[!htb]
\centering
\includegraphics[width=1.0\linewidth]{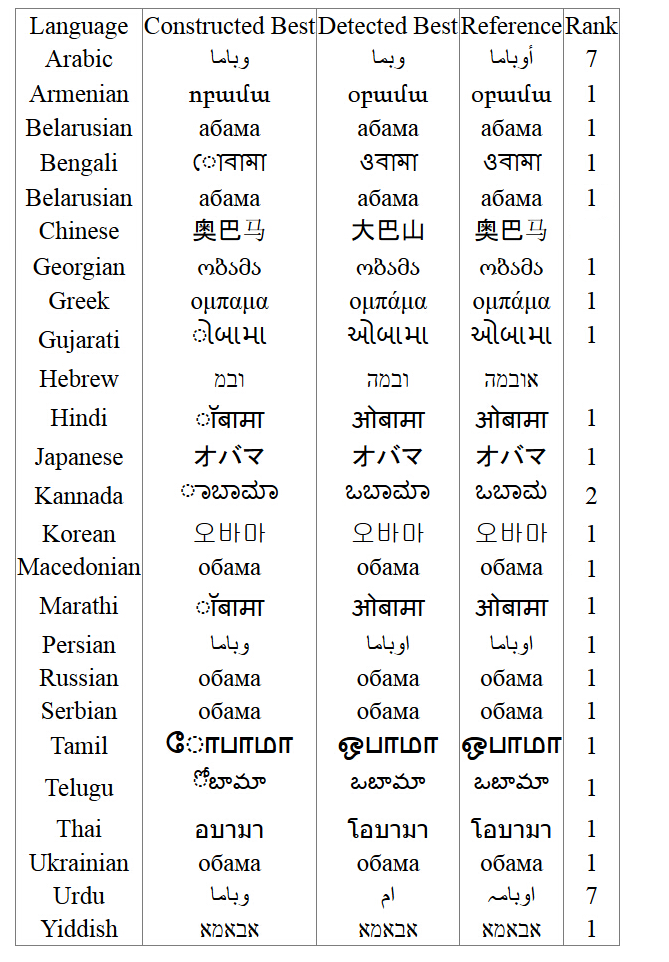}
\caption {Constructed best and detected best for word ``obama" where capitalization is disabled. Constructed best is generated via cost matrices without any prior knowledge of vocabulary. Detected best is the best match in 100,000 most frequent words in Wikipedia. Last column shows the rank of gold standard reference if it appears in these 100,000 high frequency words.}
\label{obama}
\end{figure}

In this paper, we target the problem of generating transliterations between arbitrary pairs of 69 languages, and detecting borrowed words and entities across these languages.
We train both transliteration models and semantic word embeddings in an unsupervised manner using large-scale corpus from Wikipedia.
As an example, Figure \ref{obama} shows our transliterations of the name `{\em obama} into 25 non-Latin scripts. We provide both our transliteration (constructed from scratch) and lowest edit-distance match appearing in 100,000 most frequent Wikipedia lexicons for these languages.
Our closest match proves to equal the gold standard for 20 out of 23 languages where it appears in the lexicon.
Further, our constructed best differs from the gold standard name within at most one character substitution for 22 out of 25 languages.

Our major contributions are:

\begin{itemize}
\item
{\em Training Methods for Transliteration} -- We used Wikipedia to build a training set for transliteration, starting from the cross-language links between personal and place names in Wikipedia.
We collect a dataset with 576,403 items contributing one or more transliterations from English to other languages yielding reasonable training and testing sets to learn transliterations.
But this is a very dirty training set, because many such translation pairs are not transliterations (e.g. {\em Estados Unidos} for {\em United States}).
We develop unsupervised methods to distinguish true transliterations from false, and thus clean the training set.

\item
{\em Accurate Transliteration via Substring Matching} --
We use an expectation maximization approach to use statistics of string alignments to train improved cost matrices via a Bayesian probability model.
Our methods employ substring matching instead of single-character transition matrices, enabling the recognition of phonemes, character bigrams, and beyond.

We have trained models that permit us to construct transliterations for any string between all pairs of 69 languages.
We evaluate our work against a recently-published transliteration system \cite{durrani2014integrating} which has been integrated into the Moses statistical machine translation system.
We compare our transliteration to Moses on the four languages it supports (Arabic, Chinese, Hindi, and Russian), outperforming it in 61 of 64 standards over the set of languages.



\item
{\em Distinguishing Translations from False Friends} --
Similarly spelled or sounding words can have substantially different meanings. Such pairs that span language boundaries are called false friends, e.g. {\em ropa} in Spanish means {\em clothes}, not {\em rope)}. By coupling transliteration pair analysis with semantic tests using distributed word embeddings, we can generate comprehensive lexicons of true and false friends.
Our methods get very good results in tests against human annotated standards for French (F1=0.890) and Spanish (F1=0.825).

We use our approach to generate lexicons of true and false friends between English and 69 languages. We show that the lexically-closest cohort of word pairs has a higher probability of being true friends than words that are more lexically distant in 68 out of 69 languages, indicating our methods provide a good signal to identify borrowed words.

\end{itemize}

We provide a demo that can transliterate any English string to non-Latin languages \footnote{https://soundaword.appspot.com/}.
Our transliteration code, corpora, weight matrices, and false-friend lexicons for all 69 languages will be made publicly available upon acceptance of this paper.


The rest of this paper is organized as follows. We review related work in Section \ref{related-work}. In Section \ref{data-collection}, we describe the procedure of collecting data.
Section \ref{training-model} talks about our model of learning character-based transliteration cost matrices. We analyze the performance of our model in Section \ref{results-section}.
Section \ref{friend-detection} discusses the application of detecting true and false friends.
Finally, we conclude with discussion and ideas for future work.

\section{Related Work}
\label{related-work}

Transliteration research first associates with the field of orthographic similarity detections since sound similarities co-exist with orthographic similarities \cite{brew1996word,mann2001multipath,dijkstra1999recognition,van1998orthographic,kondrak2004combining,chamizo2002false}. This work shows reasonableness of character-based transliteration between close languages (i.e. languages sharing characters) but does not discuss on distant language pairs.

Similarly, work on cognate identification also focus on close language pairs \cite{simard1993using,inkpen2005automatic,schepens2013cross,boada2013effect,kolb2008disco,kondrak2004identification,resnik2011semantic}. However, we believe multilingual transliterations contribute to even distant languages (e.g. English and Japanese) when handling OOV words and resolving ambiguities.

Further transliteration researches divide into two branches. One tries to study delicate sound changing rules of specific languages \cite{knight1998machine,abduljaleel2003statistical,suwanvisat1998thai,gao2005phoneme,virga2003transliteration,jagarlamudi2012regularized,hong2009hybrid}. Especially, an excellent ideas of using Wikipedia external links is proposed in \cite{Kirschenbaum-Wintner:2009,Kirschenbaum-Wintner:2010} and achieve promising results in English-Hebrew transliteration using Moses \cite{koehn2007moses}. However, all these systems are supervised and require extra linguistic background knowledge during processing. Plus, only one among this work evaluates transliteration on up to 4 languages and it is hard to generalize for multiple languages.


The other branch learns from only sequence of characters. One of the great advantages against sound based transliteration is that multilingual texts are much easier to obtain. \newcite{al2002machine} compares phonetic based systems with spelling based systems on transliterations between English and Arabic. \newcite{pouliquen2006multilingual} makes transliteration model based on similar spelling rules in close languages. Recent work of \newcite{durrani2014integrating} is integrated in Moses as a module, providing an unsupervised character-based transliteration training model. \newcite{matthews2007machine} proposed a proper name transliteration model on several language pairs. However, we believe utilizing character-based transliteration model can provide us with even more valuable information in natural language processing tasks. 


\section{Data Collection and Pre-processing}
\label{data-collection}



\newcite{Kirschenbaum-Wintner:2009} inspire us to use Wikipedia external links to build an aligned multilingual corpus. However, their work requires language-specific knowledge, for instance, discarding vowels and filtering out junk data using pre-defined consonant matching.
In our task, we use the names of entities (people and places) to create a training set for transliteration. \cite{francis2002impact} state that over 40\% of the brands choose to create corresponding foreign names via transliterations, By querying Freebase in categories containing 3,388,225 entries, we create a precise multilingual transliteration dictionary through Wikipedia page titles.
We then perform a rough clean up procedure to (1) unify punctuation by converting hyphens, dots, comma to underscores and, (2) remove entries which do not adhere to the (first name, last name) or whole name format Our final collection contains 576,403 English entries with multilingual mapping.


As an additional resource, we query Google translation API to get formal translations of certain English proper nouns to all 69 languages.
To reduce machine translation error, we manually pick 1,373 entities without no multi-sense ambiguities from the names of people \cite{FOSN:2000}, countries and capital cities \cite{LOCC:2014}, resulting in more than 70,000 pairs of proper name transliteration from English. Table \ref{cs} shows statistics of final data size in each language. 80\% of the final data will be used for training, 10\% is for tuning and the remaining 10\% is for testing.

\begin{table}[!htb]
\centering
\small
\begin{tabular}{|r|r||r|r|}
\hline
\multicolumn{2}{|c||}{Largest} &  \multicolumn{2}{c|}{Smallest} \\
\hline
\ Lang & Count & Lang & Count \\
\hline
\ French & 183,270 & Khmer & 1,585 \\
\ German & 178,715 & Amharic  & 2,035 \\
\ Italian & 132,545 &  Gujarati & 2,130 \\
\ Polish & 124,870 &  Maltese & 2,415 \\
\ Spanish & 107,790 & Yiddish & 2,835 \\
\ Russian & 100,085 & Kannada & 3,100 \\
\ Swedish & 91,125 & Telugu & 3,840 \\
\ Dutch & 87,870 &  Swahili & 4,620 \\
\ Portuguese & 86,515 & Haitian & 5,245 \\
\ Norwegian & 74,790 & Urdu  & 6,115 \\
\hline
\end{tabular}
\caption {Languages with the largest and smallest set of translated entities, i.e. reflecting the availability of training data.}
\label{cs}
\end{table}

\section{Training Transliteration Model}
\label{training-model}


The purpose of our training is to get a quantified measurement of sound similarities between any possible character strings in arbitrary scripts.
We expect to learn pairwise word segmentations and n-gram statistics of correlated string pieces between different languages. 

\begin{figure*}[!htb]
\centering
\includegraphics[width=1.0\linewidth]{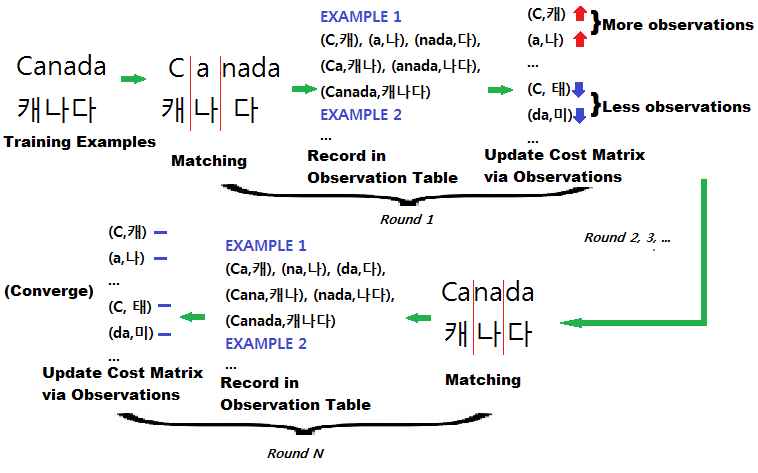}
\caption {Illustration of the training procedure. Each round we compute minimum-cost-matching and record matched string pieces for all training examples and update costs matrix through a Bayesian model.}
\label{illustration}
\end{figure*}

We maintain a cost matrix in which the cost of substituting any string $s_1$ in $Language_1$ with string $s_2$ in $Langauge_2$ will first be initialized to $len(s_1) + len(s_2)$, including empty strings. This way each training example has a fixed cost equal to the total length of two strings. After that we start an $R$ round iteration. In each round we go through all training examples and compute the minimum-cost segmentation matching. We keep tracking of all observations of matched n-grams during this round in observation table. We then adapt Bayesian setting mentioned in \cite{snyder2008unsupervised} to update cost matrix according to probability calculated by observation table. Figure \ref{illustration} illustrates the training procedure.


Data used in training example may be flawed as it might not reflect transliteration. Such training examples act as outliners during out training and we cannot find any reasonable matching even for partial string pieces. We here define ``Dirtiness" to measure how many training examples are flawed in training a specific language. Table \ref{cs2} shows 10 dirtiest among 69 languages. Big languages included in Table \ref{cs2} (e.g. Chinese, Korean) are not problematic since we have plenty of training examples. However, we expect a bad performance on small languages like Khmer and Amharic due to lack of high quality data.

\begin{table}[!htb]
\centering
\small
\begin{tabular}{|r|r||r|r|}
\hline
\multicolumn{2}{|c||}{Dirtiest} & \multicolumn{2}{c|}{Cleanest} \\
\hline
\ Lang & Dirtiness & Lang & Dirtiness \\
\hline
\ Hungarian & 41.00\% & Norwegian & 1.23\% \\
\ Amharic & 36.09\% & Bulgarian & 1.80\% \\
\ Vietnamese & 32.10\% & Macedonian & 1.83\% \\
\ Khmer & 24.58\% & Latvian & 1.99\% \\
\ Thai & 24.50\% & Russian & 2.20\% \\
\ Chinese & 23.80\% & Greek & 2.27\% \\
\ Korean & 22.20\% & Armenian & 2.41\% \\
\ Malay & 20.11\% & Georgian & 2.60\% \\
\ Tamil & 18.72\% & Czech & 2.95\% \\
\ Japanese & 16.81\% & Latin & 3.26\% \\
\hline
\end{tabular}
\caption {The 10 cleanest and dirtiest languages, defined according to the ratio of flawed examples in the training set.}
\label{cs2}
\end{table}

\begin{figure*}[!htb]
        \centering
        \begin{subfigure}[b]{0.49\textwidth}
                \includegraphics[width=\textwidth]{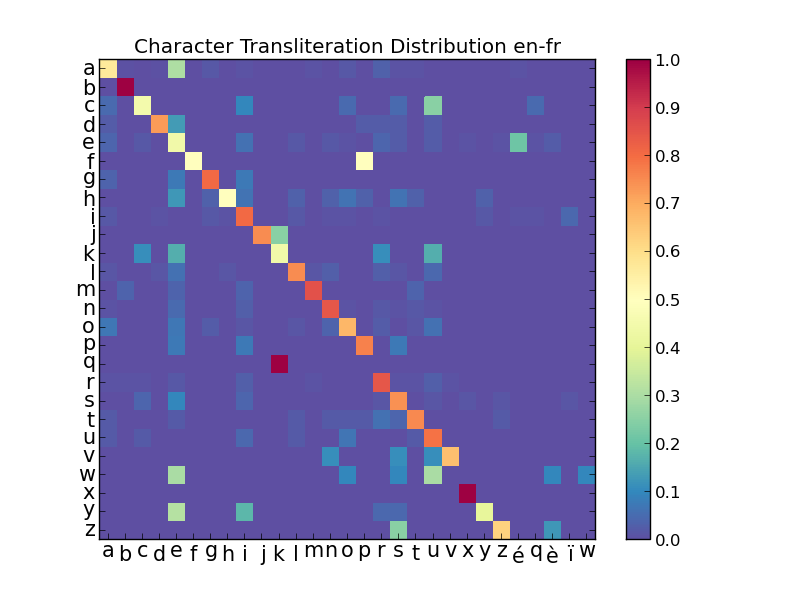}
                \caption{English-French}
                \label{heatmap_en_fr}
        \end{subfigure}
        \begin{subfigure}[b]{0.49\textwidth}
                \includegraphics[width=\textwidth]{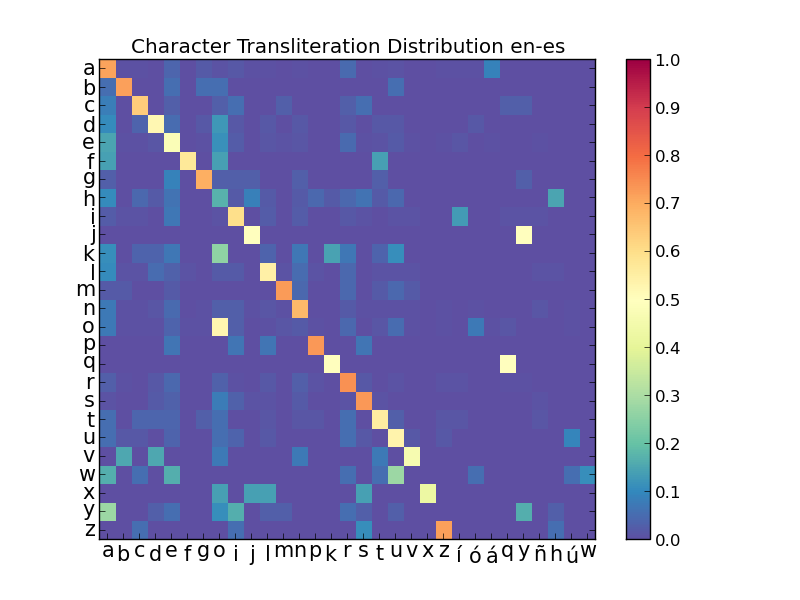}
                \caption{English-Spanish}
                \label{heatmap_en_es}
        \end{subfigure}
         \caption{Probability/cost matrix for single character pairs between English and a) French and b) Spanish. The bright diagonal shows that we discover common equivalence for most Latin characters. }
        \label{pchar}
\end{figure*}

Figure \ref{heatmap_en_fr} and Fig. \ref{heatmap_en_es} present heatmaps generated from our cost matrix showing 1-1 matching rules between characters in these languages.
The highlighted diagonals indicate strong similarity between identical Latin characters as expected, making transliteration inside the same language family meaningless. 
However, these matrices also reflect language differences: e.g.
the Spanish ``y" more often acts as an English ``j" than an English ``y";
and the Spanish ``b" is close to English ``v" while the French ``b" matches only with English ``b".

\section{Experimental Results}
\label{results-section}

In this section we evaluate the quality of our transliteration model. 

\subsection{Baseline Model}
We use the transliteration system described in \cite{durrani2014integrating} as baseline method to compare our results.
The Moses statistical machine translation system integrates their work as a module, and allows training unsupervised transliteration for OOV words.
\footnote{We use the following parameters when configuring Moses:
{\em maximum phrase length}=3,
{\em language model N-gram order}=3,
{\em language model smoothing \& interpolation}=Automatically Disabled, Interpolate;
{\em alignment heuristic}=grow-diag-final;
{\em reordering}=Monotone; and
{\em maximum distortion length}=0.
The weights for the models are:
translation model (0.2, 0.2, 0.2, 0.2, 0.2),
language model (0.5),
and distortion Model (0.0), with
word penalty=-1.
}










\subsection{Test Results}
We first compare both systems trained on our Wikipedia dataset. We focus on the performance of phrase table, i.e. measurement of sound similarities between string pieces since our dataset does not contain corpus context,
We generate the 100-best transliterations for entries in testing set on four languages of different language families: Chinese, Arabic, Hindi and Russian.

We repeated this test using third party datasets to check consistency of training models.\footnote{{\em{Chinese:}} Chinese - English Name Entity List sv1.0(LDC2005T34). 
{\em{Arabic:} } Combination of 10001 Arabic Names(LDC2005G02) and \cite{cettolo2012wit3} made available for IWSLT-13.
{\em{Hindi:} } Indian multi-parallel corpus \cite{post2012constructing}, and
{\em{Russian:} } WMT-13 data \cite{bojar2014findings}.
}

We cleaned data and retained only name mapping to feed the model, since our model does not rely on context and target a generalized method for multiple languages.
Note Moses provides several language-specific optimization methods, including weights optimizing (e.g. Mert) and Language Model Smoothing (e.g. Kneser-Ney) that might improve performance \cite{durrani2014integrating}.
However, given our goal of unsupervised transliteration, we did not attempt to employ these in our experiments.

\begin{table*}[!htb]
\begin{center}
\small
\begin{tabular}{| r | r | r | r r | r r | r r | r r | }
\hline
\ Direction & Lang & Model & \multicolumn{2}{c|}{Top-1} & \multicolumn{2}{c|}{Top-20} & \multicolumn{2}{c|}{Top-100} & \multicolumn{2}{c|}{Levenshtein 1} \\ 
\ & & & Wiki & TP & Wiki & TP & Wiki & TP & Wiki & TP \\
\hline
\multirow{4}{*}{From} & \multirow{2}{*}{ZH} & Moses & 26.7\% & 27.9\% & 44.3\% & 51.5\% & 66.1\% & 81.2\% & 64.8\% & 66.1\% \\
\ & & Ours & \bf{30.0\%} & \bf{29.8\%} & \bf{52.4\%} & \bf{53.0\%} & \bf{85.0\%} & \bf{83.3\%} & \bf{75.0\%} & \bf{79.0\%} \\
\cline{2-11}
\ & \multirow{2}{*}{AR} & Moses & 19.2\% & 20.0\% & 32.0\% & 45.0\% & 50.9\% & 80.2\% & 40.8\% & 41.6\% \\
\ & & Ours & \bf{35.2\%} & \bf{25.3\%} & \bf{60.0\%}& \bf{55.2\%} & \bf{86.3\%} & \bf{83.1\%} & \bf{60.9\%} & \bf{54.6\%} \\
\cline{2-11}
\multirow{4}{*}{EN} & \multirow{2}{*}{HI} & Moses & 23.3\% & 25.3\% & 50.4\% & 55.4\% & 70.2\% & 79.3\% & 56.1\% & 53.7\% \\
\ & & Ours & \bf{31.5\%} & \bf{30.1\%} & \bf{61.6\%} & \bf{62.5\%} & \bf{79.4\%} & \bf{83.4\%} & \bf{61.7\%} & \bf{60.3\%} \\
\cline{2-11}
\ & \multirow{2}{*}{RU} & Moses & 35.1\% & 46.1\% & 63.6\% & 69.2\% & 79.5\% & 87.5\% & 60.2\% & 70.2\% \\
\ & & Ours & \bf{40.2\%} & \bf{47.2\%} & \bf{68.1\%} & \bf{67.0\%} & \bf{82.5\%} & \bf{88.5\%} & \bf{70.5\%} & \bf{72.8\%} \\
\hline
\multirow{4}{*}{To} & \multirow{2}{*}{ZH} & Moses & 15.6\% & 21.6\% & 32.7\% & 42.1\% & 53.5\% & 73.3\% & 45.0\% & 55.0\% \\
\ & & Ours & \bf{24.1\%} & \bf{23.8\%} & \bf{51.1\%} & \bf{49.2\%} & \bf{80.4\%} & \bf{81.9\%} & \bf{60.2\%} & \bf{63.7\%} \\
\cline{2-11}
\ & \multirow{2}{*}{AR} & Moses & 20.5\% & 20.3\% & 33.9\% & 43.9\% & 49.7\%  & 79.7\% & 55.6\% & 45.6\% \\
\ & & Ours & \bf{39.0\%} & \bf{26.0\%} & \bf{77.1\%} & \bf{57.1\%} & \bf{89.3\%} & \bf{82.3\%} & \bf{75.3\%} & \bf{55.3\%} \\
\cline{2-11}
\multirow{4}{*}{EN} & \multirow{2}{*}{HI} & Moses & 21.4\% & 23.1\% & 49.8\% & 56.8\% & 71.0\% & 78.7\% & 57.3\% & \bf{62.0\%} \\
\ & & Ours & \bf{29.8\%} & \bf{29.9\%} & \bf{52.5\%} & \bf{59.7\%} & \bf{79.3\%} & \bf{79.9\%} & \bf{60.7\%} & 60.1\% \\
\cline{2-11}
\ & \multirow{2}{*}{RU} & Moses &  35.4\% & \bf{45.9\%} & 62.9\% & 70.8\% & 78.6\% & \bf{85.1\%} & 60.1\% & 70.3\% \\
\ & & Ours & \bf{39.3\%} & 44.3\% & \bf{68.7\%} & \bf{71.8\%} & \bf{81.8\%} & 84.8\% & \bf{70.0\%} & \bf{70.4\%} \\
\hline
\end{tabular}
\end{center}
\caption{Comparison of performances on Wikipedia and third party (TP) datasets. Top-k measures the percentage of correct transliterations in the top k candidates. Levenshtein 1 measures the percentage of the highest ranked transliteration that is no more than 1 substitution away from the reference transliteration, given that we consider insertion / deletion to be a special kind of substitution.}
\label{expotherdata}
\end{table*}

Figure \ref{expotherdata} shows the statistics. Our system generally outperforms Moses, winning on 61 of 64 comparisons over the eight languages and metrics.
The absolute closest transliteration (top-1) result only matches the translation target in roughly 1/3 of the test examples, indicating that there are typically a large number of transliterations of similar edit cost. Indeed, the absolute performance score substantially increases with top-20 and top-100 results, showing the need to reduce ambiguity through context matching.
Our high scores under Levenshtein 1 metric show that we generate reasonable transliteration for a large fraction of strings, retaining good lexical consistency with respect to the gold standard.
Moses’s performance substantially changes over difference training set, where we do equally well on both corpora.

\section{True and False Friends Detection}
\label{friend-detection}

\fullversion{
\begin{figure*}[!htb]
        \centering
        \begin{subfigure}[b]{0.49\textwidth}
                \includegraphics[width=\linewidth]{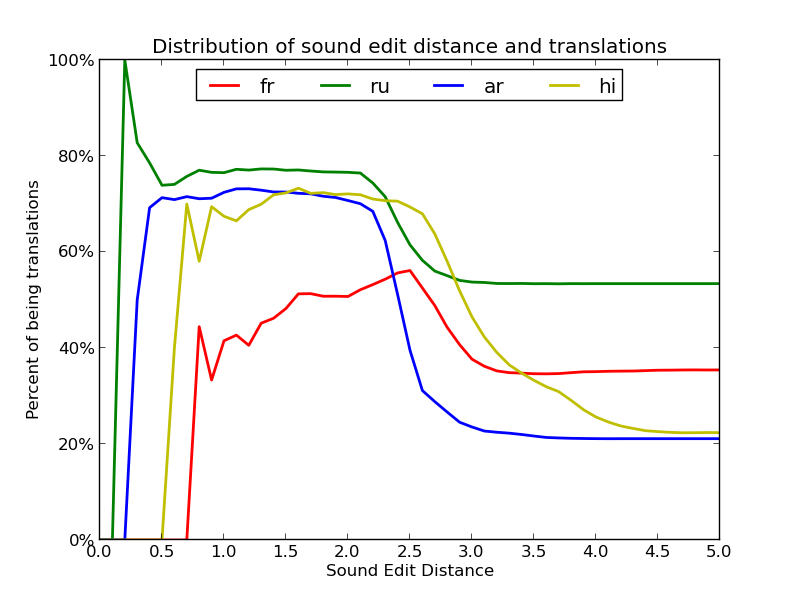}
                \caption{Precision: Percentage of finding real translations}
                \label{prec_borrowed}
        \end{subfigure}
        \begin{subfigure}[b]{0.49\textwidth}
                \includegraphics[width=\textwidth]{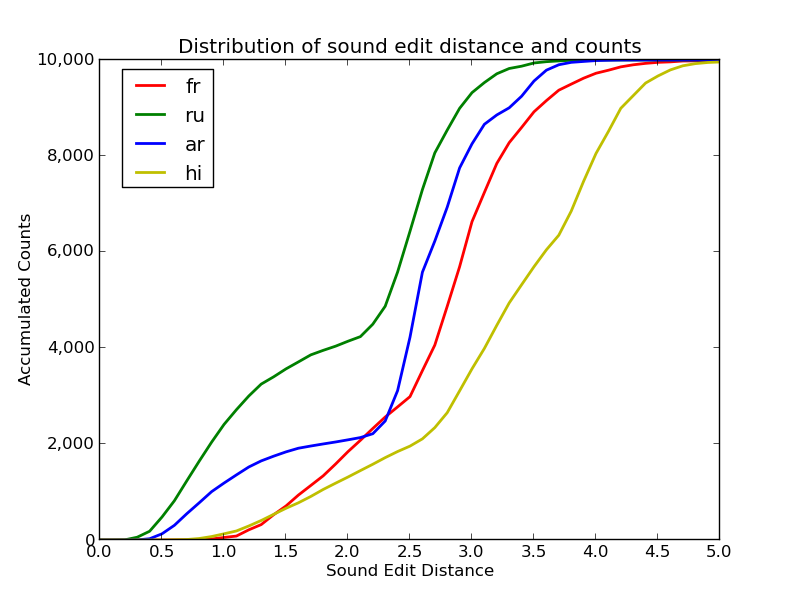}
                \caption{Recall: Accumulative counts}
                \label{rec_borrowed}
        \end{subfigure}
         \caption{Distribution of sound edit distance and the percentage of finding translations, from English to one other language.}
        \label{borrowed}
\end{figure*}
}

\begin{table*}[!htb]
\centering
\scriptsize
\begin{tabular}{|r|r|r|r|r||r|r|r|r|r||r|r|r|r|r|}
\hline
\ Lang & TP & EP & B & N & Lang & TP & EP & B & N & Lang & TP & EP & B & N \\
\hline
\ Afrikaans & 405 & 173 & 260 & 583 & Georgian & 1591 & 425 & 661 & 2434 & Nynorsk & 472 & 184 & 267 & 382 \\
\ Albanian & 639 & 295 & 533 & 1237 & \cellcolor{red}{German} & \cellcolor{red}{174} & \cellcolor{red}{640} & \cellcolor{red}{674} & \cellcolor{red}{1032} & Persian & 979 & 284 & 653 & 1753 \\
\ \cellcolor{yellow}{Amharic} & \cellcolor{yellow}{2} & \cellcolor{yellow}{0} & \cellcolor{yellow}{0} & \cellcolor{yellow}{1} & Greek & 1257 & 436 & 1042 & 1693 & \cellcolor{green}{Polish} & \cellcolor{green}{474} & \cellcolor{green}{518} & \cellcolor{green}{830} & \cellcolor{green}{1294} \\
\ \cellcolor{green}{Arabic} & \cellcolor{green}{618} & \cellcolor{green}{571} & \cellcolor{green}{922} & \cellcolor{green}{913} & Gujarati & 556 & 156 & 224 & 412 & \cellcolor{red}{Portuguese} & \cellcolor{red}{283} & \cellcolor{red}{660} & \cellcolor{red}{1150} & \cellcolor{red}{567} \\
\ Armenian & 1141 & 405 & 449 & 1003 & Haitian & 220 & 235 & 76 & 75 & \cellcolor{green}{Romanian} & \cellcolor{green}{388} & \cellcolor{green}{618} & \cellcolor{green}{1071} & \cellcolor{green}{888} \\
\hline
\ Azerbaijani & 854 & 166 & 284 & 1244 & \cellcolor{green}{Hebrew} & \cellcolor{green}{670} & \cellcolor{green}{466} & \cellcolor{green}{864} & \cellcolor{green}{1248} & \cellcolor{red}{Russian} & \cellcolor{red}{854} & \cellcolor{red}{1143} & \cellcolor{red}{3344} & \cellcolor{red}{1202} \\
\ Basque & 352 & 184 & 187 & 597 & Hindi & 1187 & 323 & 725 & 1043 & \cellcolor{green}{Serbian} & \cellcolor{green}{1695} & \cellcolor{green}{743} & \cellcolor{green}{2022} & \cellcolor{green}{1027} \\
\ Belarusian & 1561 & 291 & 765 & 1295 & \cellcolor{green}{Hungarian} & \cellcolor{green}{227} & \cellcolor{green}{141} & \cellcolor{green}{244} & \cellcolor{green}{2762} & \cellcolor{green}{Serbo-Croat} & \cellcolor{green}{546} & \cellcolor{green}{474} & \cellcolor{green}{715} & \cellcolor{green}{1442} \\
\ Bengali & 661 & 227 & 280 & 1310 & Icelandic & 349 & 82 & 95 & 1156 & \cellcolor{green}{Slovak} & \cellcolor{green}{623} & \cellcolor{green}{310} & \cellcolor{green}{696} & \cellcolor{green}{521} \\
\ Bosnian & 760 & 282 & 389 & 1392 & \cellcolor{green}{Indonesian} & \cellcolor{green}{198} & \cellcolor{green}{231} & \cellcolor{green}{478} & \cellcolor{green}{1802} & \cellcolor{green}{Slovenian} & \cellcolor{green}{518} & \cellcolor{green}{296} & \cellcolor{green}{552} & \cellcolor{green}{632} \\
\hline
\ \cellcolor{green}{Bulgarian} & \cellcolor{green}{2231} & \cellcolor{green}{674} & \cellcolor{green}{2599} & \cellcolor{green}{1562} & Irish & 152 & 155 & 57 & 837 & \cellcolor{red}{Spanish} & \cellcolor{red}{530} & \cellcolor{red}{788} & \cellcolor{red}{1640} & \cellcolor{red}{1115} \\
\ \cellcolor{green}{Catalan} & \cellcolor{green}{639} & \cellcolor{green}{733} & \cellcolor{green}{1255} & \cellcolor{green}{1187} & \cellcolor{red}{Italian} & \cellcolor{red}{300} & \cellcolor{red}{613} & \cellcolor{red}{1272} & \cellcolor{red}{581} & Swahili & 113 & 96 & 70 & 458 \\
\ \cellcolor{yellow}{Khmer} & \cellcolor{yellow}{21} & \cellcolor{yellow}{194} & \cellcolor{yellow}{1} & \cellcolor{yellow}{41} & \cellcolor{green}{Japanese} & \cellcolor{green}{1048} & \cellcolor{green}{414} & \cellcolor{green}{1112} & \cellcolor{green}{854} & \cellcolor{green}{Swedish} & \cellcolor{green}{199} & \cellcolor{green}{446} & \cellcolor{green}{564} & \cellcolor{green}{461} \\
\ \cellcolor{green}{Chinese} & \cellcolor{green}{18} & \cellcolor{green}{142} & \cellcolor{green}{26} & \cellcolor{green}{0} & Kannada & 737 & 201 & 330 & 1173 & Tagalog & 211 & 118 & 161 & 587 \\
\ \cellcolor{green}{Croatian} & \cellcolor{green}{446} & \cellcolor{green}{522} & \cellcolor{green}{803} & \cellcolor{green}{984} & \cellcolor{green}{Korean} & \cellcolor{green}{150} & \cellcolor{green}{488} & \cellcolor{green}{307} & \cellcolor{green}{81} & Tamil & 386 & 189 & 239 & 1121 \\
\hline
\ \cellcolor{green}{Czech} & \cellcolor{green}{405} & \cellcolor{green}{699} & \cellcolor{green}{1142} & \cellcolor{green}{387} & Latin & 352 & 346 & 158 & 1184 & Telugu & 821 & 231 & 399 & 1322 \\
\ \cellcolor{green}{Danish} & \cellcolor{green}{251} & \cellcolor{green}{396} & \cellcolor{green}{493} & \cellcolor{green}{403} & Latvian & 812 & 376 & 491 & 2188 & \cellcolor{yellow}{Thai} & \cellcolor{yellow}{267} & \cellcolor{yellow}{109} & \cellcolor{yellow}{65} & \cellcolor{yellow}{799} \\
\ \cellcolor{green}{Dutch} & \cellcolor{green}{208} & \cellcolor{green}{431} & \cellcolor{green}{360} & \cellcolor{green}{449} & Lithuanian & 489 & 542 & 473 & 505 & \cellcolor{green}{Turkish} & \cellcolor{green}{263} & \cellcolor{green}{350} & \cellcolor{green}{534} & \cellcolor{green}{1082} \\
\ \cellcolor{green}{Esperanto} & \cellcolor{green}{522} & \cellcolor{green}{410} & \cellcolor{green}{638} & \cellcolor{green}{750} & Macedonian & 2926 & 422 & 1769 & 1667 & \cellcolor{green}{Ukrainian} & \cellcolor{green}{1718} & \cellcolor{green}{781} & \cellcolor{green}{3046} & \cellcolor{green}{1361} \\
\ Estonian & 245 & 161 & 150 & 605 & \cellcolor{green}{Malay} & \cellcolor{green}{226} & \cellcolor{green}{140} & \cellcolor{green}{263} & \cellcolor{green}{1465} & Urdu & 338 & 111 & 146 & 879 \\
\hline
\ \cellcolor{green}{Finnish} & \cellcolor{green}{157} & \cellcolor{green}{285} & \cellcolor{green}{159} & \cellcolor{green}{605} & Maltese & 149 & 214 & 84 & 88 & \cellcolor{yellow}{Vietnamese} & \cellcolor{yellow}{46} & \cellcolor{yellow}{20} & \cellcolor{yellow}{10} & \cellcolor{yellow}{2976} \\
\ \cellcolor{green}{French} & \cellcolor{green}{486} & \cellcolor{green}{949} & \cellcolor{green}{2108} & \cellcolor{green}{1804} & Marathi & 410 & 181 & 225 & 751 & Welsh & 248 & 81 & 107 & 992 \\
\ \cellcolor{green}{Galician} & \cellcolor{green}{664} & \cellcolor{green}{511} & \cellcolor{green}{934} & \cellcolor{green}{1343} & \cellcolor{green}{Norwegian} & \cellcolor{green}{216} & \cellcolor{green}{414} & \cellcolor{green}{486} & \cellcolor{green}{401} & \cellcolor{yellow}{Yiddish} & \cellcolor{yellow}{626} & \cellcolor{yellow}{206} & \cellcolor{yellow}{156} & \cellcolor{yellow}{803} \\

\hline
\end{tabular}
\caption {Statistics of True and False Friends between English and all 69 languages.
TP denotes Google translation pairs which are not close in our word embeddings.
EP denotes close embedding pairs not recognized as translations by Google, while
B denotes words pairs passing both semantic tests,
N denotes false friends: word pairs which pass neither semantic test.
Red languages are those with the highest ration of B / TP, showing a significant correlation between our embedding and Google translation.
At least 50\% of Google translation pairs survives our embedding test for all 33 Red or Green languages. By contrast, the yellow languages are those where the embedding test performed poorly.}
\label{truefriends}
\end{table*}

Although our transliteration model is accurate at detecting lexical similarity across languages, words that look alike or sound alike do not necessarily mean the same thing.
{\em False friends} are word pairs across languages that look the same, but mean something different. For example, the Spanish word {\em ropa} means clothes, not rope.

Such false friends are the bane of students learning foreign languages.
For our transliteration tests to identify true language borrowings, we must also establish that the words have similar semantics.
To perform such a test, we relied on the Polyglot distributed word embeddings presented in \cite{al2013polyglot}.
The $L_2$ norm between two word representations captures its semantic distance.

\begin{figure}[!htb]
\centering
\includegraphics[width=1.0\linewidth]{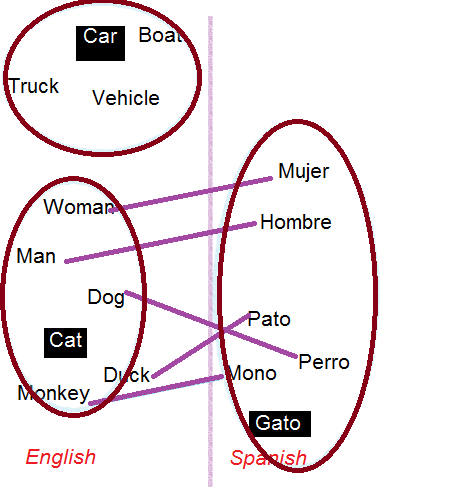}
\caption {Illustration of word embedding test. In case no direct evidences of semantic similarity between ``Cat" and ``Gato" are found, we check number of translations that links nearest neighbors of ``Cat" and ``Gato" . Since  ( ``Dog", ``Monkey", ``Duck") matches perfectly with (``Perro", ``Mono", ``Pato"), we can judge that (``Cat", ``Gato" ) has close semantic meanings. (``Car" , ``Gato")  will definitely fail this test.}
\label{cartoon2}
\end{figure}

However, the Polyglot embeddings do not reside in same geometric space of latent dimensions for different languages. Thus instead of directly computing the distance between representations across languages, we check how many pairs of known translations lie within the 300 words closest words in each language in case we are lack of direct translation evidences. This process is illustrated in Figure \ref{cartoon2}.

\swallow{
Though our model focuses on generating transliteration from segmentations, it is still worth checking if our cost matrix can detect word pairs with similar pronunciations in the dictionary. We pick words with 5 or more characters from most frequent 100,000 lexicons in Wikipedia described in \cite{al2013polyglot}. We then calculate pairwise alignment cost according to our final cost matrix for difference and record 10,000 pairs with lowest cost. These word pairs will then be tested against collected Google translation \cite{chen2014building}. Plus, in order to reduce MT error, we also check if there are strong translation signals between closest neighbors in each other’s embedding space since word representation has the feature of grouping words with similar syntactic / semantic behavior. Illustration of this extra embedding test is shown in Fig. \ref{cartoon2}. 
}

\subsection{Evaluation against Human Annotation}

For two languages (French and Spanish) we found published lists of true and false friends with English.
We did an evaluation of our results against these human-annotated gold standards, in particular
1756 French-English cognates and 541 false friends suggested in \cite{inkpen2005automatic} as well as 1345 of Spanish-English cognates \footnote{http://spanishcognates.org/} plus 217 false friends \footnote{http://www.esdict.com/}.
Our performance is shown in the table below:


\begin{center}
\small
\begin{tabular}{|r|r|r|}
\hline
\ Lang & F1 & Acc \\
\hline
\ French &  0.890 & 89.2\% \\
\ Spanish & 0.825 & 82.3\% \\
\hline
\end{tabular}
\end{center}

Our methods yield substantial agreement with these published standards, demonstrating the general soundness of our approach.

\fullversion{
Fig. \ref{borrowed} shows statistics of 4 languages mentioned in previous tests, including French, Russian, Arabic and Hindi. Chinese is not included since most Chinese words are less than 3 characters and are not picked in the first step. There is a drop near distance of 2.5 for all these four languages and best F1-score are picked close to this point. Since initial value of substitution between two non-related characters is set to 2, such phenomenon indicate a strong signal of ``less likely to be share sound similarity". Threshold of 2.5 help group 20\% of the word pairs into a high-similarity group while remaining word pairs are considered to be in low-similarity group, though they are much closer than random word pairs.Table \ref{truefriends} shows statistics of true-friends and false friends we detected, plus p-value showing significant differences between behaviors of words with low sound similarities and those with high sound similarities:
}

\begin{figure*}[!htb]
\centering
\includegraphics[width=0.45\linewidth]{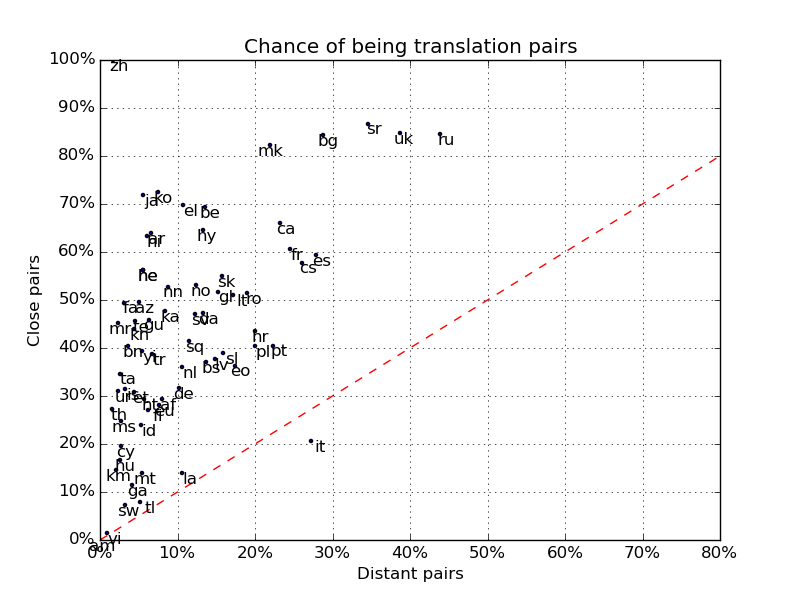}
\hspace{0.25in}
\includegraphics[width=0.45\linewidth]{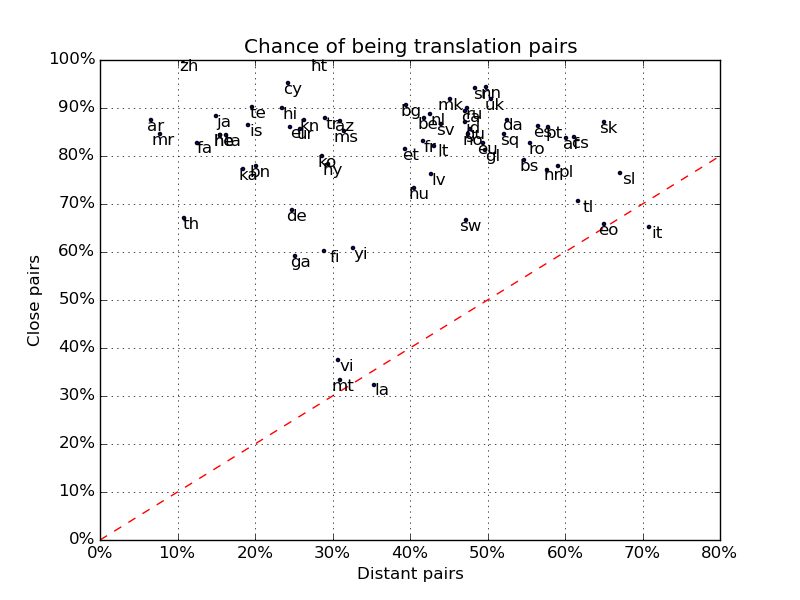}
\caption{(left) Fraction of gold standard translations within very close edit distance pairs ($d < 2$) versus the next closest 10,000 pairs. (right) Same fractions after retaining only the 50\% of pairs which are closest by embedding distance.
For 68 of 69 languages, the lexically closer pairs are more likely to be translations (left).
Further, eliminating pairs failing the embedding test shifts all languages to the upper right, showing that the embedding test accurately captures semantic similarity (right).}
\label{plot}
\end{figure*}

\subsection{Cross-Language Scan}

Emboldened by these results, we performed a search for lexically/semantically similar words between English and all 69 of our transliterated languages.
The results appear in Table \ref{truefriends}.
For each language, we report the number of false friends we identify (column N).
The other three columns reflect different notions of true friends: single-word translations according to Google (TP), near neighbors in embedding test (EP), and those which survive both of these semantic tests (B).

Without a language-specific analysis of each of the classes, it is difficult to determine which of these reflect language borrowings most accurately.
The quality of the word embeddings vary substantially by language, as does the quality of Google's translation support.
Our preferred measure of quality is the ratio of word pairs which survive both tests (B) over all that having Google translations (B+TP).
The 33 languages colored red and green all have a ratio of $>0.5$, indicating the highest quality embeddings.
The red languages denote the five with the best embeddings, with the poorest five (in yellow) reflect languages with excessively small training data (Amharic and Khmer).
Our methods have a particularly difficult time with Vietnamese, which bases a misleading similarity to Latin languages at the character level.

\subsection{Cross-Language Validation}

Figure \ref{plot} provides a deeper assessment of our cross-language scan.
For each language, we identified which words in its 100,000 word vocabulary were lexically very similar (edit distance $\leq 2$, which is decided by initial value of substitution) to a word in the English vocabulary.
We then considered the next closest 10,000 word pairs, which should also be enriched in real transliterations (by contrast, only 0.01\% random word pairs have a translation link) -- but less enriched than the initial cohort.
Indeed, Figure \ref{plot} (left) shows this to be true for 68 of 69 languages, denoted by points in the upper left triangle.

To establish that our embedding test accurately eliminates false friends, we pruned the lower half of each cohort according to the embedding test, i.e. retained only those words whose distance in embedding space was below the median value.
Figure \ref{plot} (right) shows that this action dramatically shifts each language up and to the right. With the exception of three outlier languages (Vietnamese, Latin, and Maltese), well over 50\% of our closest cohort are now true friends (translations).
For somewhat more than half of the languages, the lexicographically second cohort is now rich in true friends to the 50\% level.

\swallow{
Fig. \ref{plot} shows a huge boost in finding translations in high-similarity group. The probability of find translation is more than doubled compared with low-similarity group, indicating our cost matrix measure sound consistency well. There are outliners however, including languages with small and dirty datasets (e.g. Amharic, Khmer), languages with short words (e.g. Chinese) and languages with lots of inflections (e.g. Italian) that increases the final cost of alignment.
}

\section{Conclusion}

In this paper, we have developed transliteration models that accurately identify borrowed out of vocabulary (OOV) words, for 69 different languages.
We have evaluated our transliterations against published gold standards when available and against intrinsic measures when such standards are not available.
Further, we demonstrated that adding word embeddings to provide a semantic test enables us to distinguish true borrowings from false friends.

There are several directions to improve the future quality of our transliteration model:
\begin{itemize}
\item
{\em Phonetic Information} -- Our models improve with additional training data, particularly for resource-poor languages.
An exciting way to increase this volume would be aligning speech translations as represented in a phonetic dictionary or sound system (e.g. IPA) as suggested in \cite{jagarlamudi2012regularized}.


\item
{\em Multiple transliteration} -- Though English Wikipedia has the richest resources in the world,
it is not guaranteed that English is the source language of borrowed names.
Currently we employ a star network of transliteration pairs centered through English.
A richer graph with other important languages (e.g. Russian and Chinese) would improve performance.

\item
{\em Longer-Range Dependencies} – As we target transliteration, our model should utilize longer range dependencies.
Observe that a silent e at the end of English words changes the pronunciation of vowels earlier in the word, so the ``li" is different in ``lit" and ``like". Under context the Moses system with optimization exploits such phenomena, but we believe with we can learn such pronunciation features from the text itself.

\end{itemize}

\section*{Acknowledgments}
This research was partially supported by NSF Grants DBI-1060572 and IIS-1017181, and a Google Faculty Research Award.

\end{document}